%% file: main.tex
\documentclass{article}  
\usepackage[margin=1.1in]{geometry}

\usepackage{subcaption}
\usepackage{csquotes}
\usepackage[labelfont=bf]{caption}
\usepackage{textcomp}

\usepackage[T1]{fontenc}
\usepackage[utf8]{inputenc}
\usepackage[british]{babel}
\usepackage[table]{xcolor}
\usepackage{collcell}
\usepackage{hhline}
\usepackage{pgf}
\usepackage{multirow}

\def\colorModel{hsb} 

\newcommand\ColCell[1]{
  \pgfmathparse{#1<50?1:0}  
    \ifnum\pgfmathresult=0\relax\color{white}\fi
  \pgfmathsetmacro\compA{.52}   
\pgfmathsetmacro\compB{#1/10} 
\pgfmathsetmacro\compC{1-#1/100}        
  \edef\x{\noexpand\centering\noexpand\cellcolor[\colorModel]{\compA,\compB,\compC}}\x #1
  } 
  
\newcolumntype{E}{>{\collectcell\ColCell}m{0.3cm}<{\endcollectcell}}  
\newcommand*\rot{\rotatebox{90}}

\usepackage[colorinlistoftodos]{todonotes}

\title{\LARGE \bf By Land, Air, or Sea: \\Multi-Domain Robot Communication Via Motion}

\author{Michael Fulton$^{1}$, Mustaf Ahmed$^{2}$, and Junaed Sattar$^{3}$
\thanks{*This work was supported by the MnDRIVE Initiative.}
\thanks{The authors are from the University of Minnesota, Twin Cities, Minneapolis, MN, USA.
        {\tt\small \{$^{1}$fulto081, $^{2}$ahmed719, $^{3}$junaed\}@umn.edu}}%
}

\begin{document}

\maketitle
\thispagestyle{empty}
\pagestyle{empty}

\input{src/abstract.tex}
\input{src/introduction.tex}

\input{src/related.tex}
\input{src/problem.tex}
\input{src/implementation.tex}
\input{src/experimentalsetup.tex}
\input{src/results.tex}
\input{src/conclusion.tex}
\input{src/acknowledgment.tex}
\clearpage

\bibliographystyle{abbrv}
\bibliography{RCVM.bib}

\end{document}

%% file: src/abstract.tex
\begin{abstract}
In this paper, we explore the use of motion for robot-to-human communication on three robotic platforms: the 5 degrees-of-freedom (DOF) Aqua autonomous underwater vehicle (AUV), a 3-DOF camera gimbal mounted on a Matrice 100 drone, and a 3-DOF Turtlebot2 terrestrial robot.
While we previously explored the use of body language-like motion (called kinemes) versus other methods of communication for the Aqua AUV, we now extend those concepts to robots in two new and different domains. 
We evaluate all three platforms using a small interaction study where participants use gestures to communicate with the robot, receive information from the robot via kinemes, and then take actions based on the information.  
To compare the three domains we consider the accuracy of these interactions, the time it takes to complete them, and how confident users feel in the success of their interactions.
The kineme systems perform with reasonable accuracy for all robots and experience gained in this study is used to form a set of prescriptions for further development of kineme systems.
\end{abstract}

%% file: src/introduction.tex
\section{INTRODUCTION}
Through the history of robotics, methods of robot control and communication have slowly shifted and changed.
While most early robots required expert knowledge to control through the use of a terminal~\cite{nilsson_shakey_1984}, communicating with modern robots has grown ever easier.  
For example, Roomba\texttrademark{} iSeries robots can be controlled with the iRobot HOME smartphone application and social robots such as Pepper~\cite{pandey_mass-produced_2018} use a combination of computer vision, speech synthesis, and digital displays to create a more natural human interface.
Indeed, the pursuit of natural, effective, and well-integrated human interfaces for robotics has been a topic at the center of human-robot interaction (HRI) research since the inception of the field.
The more similar human-robot interaction is to human-human interaction, the easier it will be for robots to take a place in everyday society as co-workers and aides, helping in aspects of life such as healthcare, home care, and most occupations.

For this reason, we choose to explore the use of motion as a method of communication from a robot to a human in interaction scenarios with robots from three different domains: land, air, and water (seen in Figure \ref{fig:robbits}).
The use of motion as a method of communication is attractive because humans comprehend the information expressed in each other's motion naturally and without training~\cite{wood_children_1976}. 
We have a shared social language of motion comprised of implicit body language and explicit gestures which we learn from an early age.
If we can effectively tap into that shared language, motion-based communication for robots could be as natural as the motion communication we use with each other. 
This would allow robots to communicate without having to maintain visibility of a digital display which can have strict viewing angles, without having to use speech synthesis which can be disruptive or inaudible depending on the environment, and without using peripheral devices which add unnecessary complexity to interactions.

This idea has been explored thoroughly in the case of humanoid robots, which have an extensive body of work exploring facial expression synthesis~\cite{breazeal_emotion_2003} and body language~\cite{brooks_behavioral_2007}.
These methods are often focused on affective (emotional) expression and take advantage of the humanoid form of their robots. 
A more limited set of work has explored motion-based communication in non-humanoid robots~\cite{bethel_robots_nodate}, drones~\cite{arroyo_daedalus:_2014, duncan_investigation_2018}, and x-y-theta terrestrial robots~\cite{knight_expressive_2014}.
This work is also often focused on the topic of affective expression, rather than the informative communication that is the topic of this work.

\input{figures/robbits.tex}
In this work, we focus on information communication via motion using a method we have previously proposed~\cite{fulton_robot_2018}: \enquote{body language} gestures called kinemes~\cite{noth_handbook_of_semiotics_1995}.
We bring our work with kinemes, which had previously focused on underwater robots and only been tested in simulations, to physical platforms and perform tests in the real world.
Our implementations are on  robots from the three domains we specified: aquatic, aerial, and terrestrial robots.
To evaluate the effectiveness of these kinemes across our chosen domains, we conduct a small study comparing the accuracy, interaction times, and user confidence in the interaction.
However, we do not compare kineme communication to traditional communication vectors in each field, a larger question which we intend to address in future work.
Our results reveal interesting aspects of kineme communication and a promising future for the kineme systems in all three domains of interaction, despite the significant differences in the motion capabilities of the different robotic platforms.

\subsection*{\textbf{Contributions}}
The contributions of this work are the following:
\begin{itemize}
    \item An implementation of our previously proposed kineme communication system for the Aqua AUV.
    \item Implementations of kineme communications systems for the Matrice 100 drone and the Turtlebot2 terrestrial robot.
    \item Results from an interaction study on all three platforms, including accuracy and speed of interaction sequences.
    \item Discussion of said results, including avenues for further research.
\end{itemize}

%% file: figures/robbits.tex
\begin{figure}
    \centering
    \includegraphics[width=0.8\textwidth]{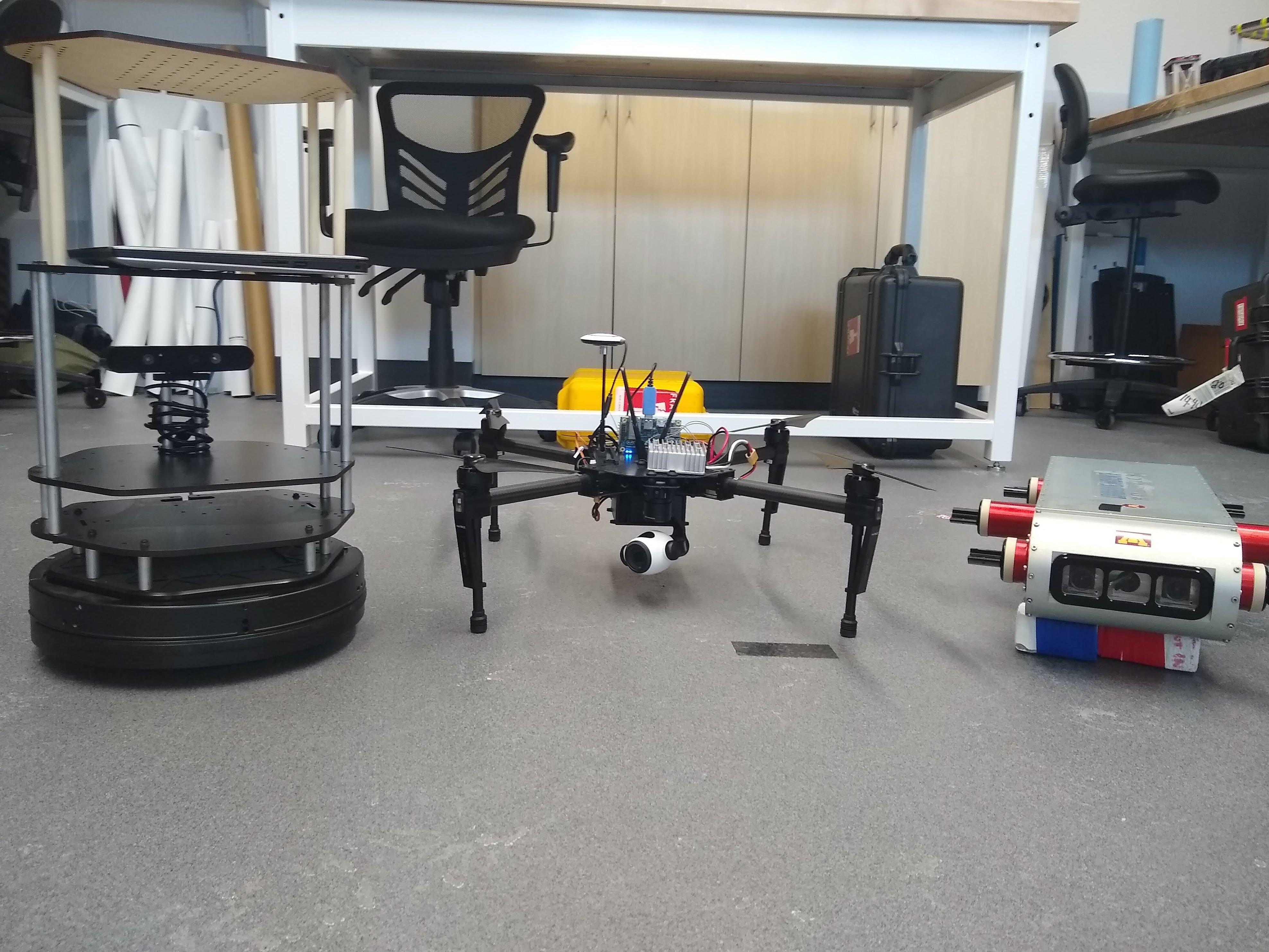}
    \caption{The three robots used in this study. From left to right: Turtlebot2, Matrice 100 with Z3 gimbal, Aqua.}
    \label{fig:robbits}
    \vspace{-7mm}
\end{figure}

%% file: src/related.tex
\section{RELATED WORK}
While a great deal of previous work in human-robot interaction is relevant to this topic, we can generally separate relevant works in motion-based or other nonverbal human-to-robot communication methods by the robot type for which the interaction is designed: underwater, aerial, or terrestrial. 
With terrestrial robots, we make the distinction that the terrestrial robots are non-humanoid, as underwater robots and aerial robots usually have no human features, whereas terrestrial robots vary significantly. 
While there is interesting work on the topic of humanoid, motion-based HRI~\cite{breazeal_emotion_2003, mavridis_review_2015}, it is not particularly relevant to our problem because humanoid and non-humanoid motion communication have different methodologies and issues.

\subsection{Underwater Robots}
Underwater HRI is a difficult topic due to the degradation of most common communication vectors including poor visibility, distorted audio, and attenuated RF signals.
When specifically considering robot-to-human communication, the most common method is the use of small digital displays~\cite{sattar_enabling_2008, sattar_ensuring_2014,islam_dynamic_2018}. 
A prominent example of this type of interactive setup is CADDY, the Cognitive Autonomous Diving Buddy, which uses a display to communicate with divers, who in turn send the robot commands using hand gestures~\cite{chavez_robust_2018}.
Hand-gesture based human-to-robot communication is a common technique~\cite{demarco_underwater_2014, islam_dynamic_2018} for underwater HRI, although other techniques have been used in the past, such as using fiducial markers as \enquote{flash cards}~\cite{dudek_visual_2007}.
Another example of digital displays being used by underwater robots is Swimoid~\cite{ukai_swimoid:_2013}, which uses a large display to interact with a diver.
Purpose-built interaction devices used for bidirectional communication are also common, as in the work of Verzjilenberg et.al.~\cite{verzijlenberg_tablet_2010}. 
One of the more unique proposals is the use of an AUV's in-built light system in Demarco et.al.~\cite{demarco_underwater_2014}, where variations in illumination intensities are used to communicate a small set of concepts. 
The study in that work was one of the first to show that humans and robots could collaborate on complex tasks underwater without the use of digital displays or dedicated interaction devices.
Work from Murphey et.al.~\cite{murphy-chutorian_head_2009} also explored the problem of interaction with aquatic robots, mostly surface vehicles, in the context of a bridge inspection task.

\subsection{Aerial Robots}
In the world of aerial drones, a prominent example of motion-based communication is the Daedelus social unmanned aerial vehicle (s-UAV)\cite{arroyo_daedalus:_2014}, which had a motorized \enquote{head} with colored eyes.
Daedelus focused on expressing emotions to interactants, through a combination of head motions, eye colors, and propeller states.
The work of Cauchard et. al.~\cite{cauchard_emotion_2016} also explores motion as a method of communication.
Their work concerned emotional display through modifications of flight commands by varying trajectories and speeds.
More recently, Duncan et.al.~\cite{duncan_investigation_2018} used flight motion to communicate information to users and achieved great success. 
We do not utilize the flight motion of the Matrice, instead focusing on the camera gimbal attached to our drone for our kinemes. 
This has previously been explored by the use of a pan-tilt camera in Wainer et.al.~\cite{wainer_role_2006} to simulate head gestures by controlling the angle of the camera.
This pan-tilt system is designed to instruct participants in playing a disk stacking game called the Towers of Hanoi.
Compared to this method, our work supports more concepts in its communication and is more general in the types of interaction scenarios it could support.
Of course, a significant number of works have explored the human-to-robot communication side of HRI, mostly focusing on the use of gestures to control drone action~\cite{ng_collocated_2011, monajjemi_hri_2013}, sometimes in combination with facial pose or gaze estimates~\cite{nagi_human_2014, hansen_use_2014}.
However, the majority of work with aerial robots uses wireless communication between computers to manage interactions between the robot and its operators~\cite{beachly_uas-rx_2017}. 
This allows for complex control and information delivery, with the only drawbacks being the requirement of wireless communication and the addition of a dedicated control device, which could impede the mobility of the human interactants. 

\subsection{Terrestrial Robots}
Terrestrial robots exist in many shapes and sizes, and non-humanoid terrestrial robots have a wide variety of HRI work in their history.
As previously mentioned, much motion-related HRI focuses on terrestrial humanoid robots, much of it concerned with affective display. 
There are a number of works which directly relate to our problem of robot-to-human communication using non-verbal, non-facial methods with a non-humanoid terrestrial robot. 
Bethel's~\cite{bethel_survey_2008} doctoral dissertation~\cite{bethel_robots_nodate} is a seminal work in this field, exploring the use of position, orientation, motion, colored lights, and sound to add affective display to appearance-constrained robots used for search and rescue. 
Similar communication has also been applied to a canine-like robot ~\cite{moshkina_human_2005} and learned through an interactive, evolutionary process based on human feedback~\cite{shimokawa_acquiring_2001}. 
While many of these works focus on the actual display of the emotions and less on how they are generated, Novikova et al.~\cite{novikova_design_2014} models a generative emotional framework which displays generated robot emotions through motion.
Affective display HRI does not completely encompass non-verbal, non-facial HRI, however.
One example is Baraka et al.~\cite{baraka_mobile_2018}, in which state information such as the intended path is displayed using lights distributed around the robot's body. 
One of the most interesting works in this space is that of Knight et al~\cite{knight_expressive_2014}, which uses the motion of an x-y-theta robot to display the internal state and task state of the robot. 
This is relatively similar to our Turtlebot2 implementation of kinemes but focuses more on affective and state display rather than actual interaction.

%% file: src/problem.tex
\section{PROBLEM DEFINITION}
In previous work~\cite{fulton_robot_2018}, we established the viability of body language motions (kinemes) as a method of robot-to-human communication for underwater robots.
However, beyond the limitation to underwater robots, this work used only simulated or animated kinemes.
In this work, we seek to expand the use of kinemes to multiple robotic platforms, exploring the comparative performance of this method in different domains. 
We also expand on previous work by physically implementing kinemes for all the robotic platforms we test.

\subsection{Pilot Study Findings}
Our pilot study from previous work found reason to further explore kinemes as a method of robot-to-human communication, by discovering that kinemes could outperform a flashing light-based system of communication in accuracy. 
While the kineme system required training for users to become as quick at communicating as with the light system, no training was required for improved accuracy. 
Regardless of whether they were given no explanation of the training system or a full training, users still performed better with the kineme system overall and improved in their recognition accuracy and speed as their level of training improved.
These results established the validity of kinemes as a method of information communication which could be learned effectively.

\subsection{Goals of This Work}
In this work we explore three aspects of kineme communication that have not previously been explored:

\subsubsection{Physical Implementation}
In this work we deal with the first implementation of our kinemes on physical platforms. 
This greatly increases the overhead required to develop kinemes for study, as physical implementations require much more testing and development compared to the simulations we used previously.
However, this also gives us the advantage of enhanced realism over a simulated robot, which may be helpful for users properly understand the kineme.

\subsubsection{Different Domains}
We explore our kinemes for the first time in different domains besides underwater. 
This adds other considerations to the development of kinemes in terms of the types of motions possible.
For instance, an aerial robot cannot generally do an overhead loop, while an AUV can do so easily. 
Additionally, by the nature of the fluid medium in which they move, aerial drones are somewhat more \enquote*{lurchy} while marine robots begin and end their movements slower because of the resistance of water.

\subsubsection{Levels of Anthropomorphism}
Since multiple robots are being used in this work, we can also explore different types of anthropomorphism. 
Anthropomorphism or \enquote*{humanization} was a core concept behind the development of the original kinemes, attempting to lead users to use their human body language intuition to interpret kinemes.
In the case of these robots, different body features may lead to different anthropomorphizations, which will add another interesting dimension to the understanding of kinemes.

\input{figures/kineme_list.tex}

%% file: figures/kineme_list.tex
\begin{table}[]
    \vspace{2mm}
    \centering
    \begin{tabular}{r|l}
         \textbf{Kineme} & \textbf{Meaning} \\ \hline
         \textit{Affirmative} & Yes, affirmative\\
         \textit{Negative} & No, negative\\
         \textit{Follow Me} & Follow the robot\\
         \textit{Indicate Movement} & Move in the direction the robot indicates\\
         \textit{Indicate Object} & Pick up/look at the object the robot indicates\\
         \textit{Indicate Stay} & Remain in the current location\\
         \textit{Danger} & Retreat from area, danger here.\\
         \textit{Malfunction} & Some robot system is malfunctioning\\
         \textit{Repeat Last} & Repeat the previous command
    \end{tabular}
    \caption{The nine kinemes used in this work.}
    \vspace{-7mm}
    \label{tab:kineme_list}
\end{table}

%% file: src/implementation.tex
\section{Implementation}
For this work, nine kinemes (listed in Table \ref{tab:kineme_list}) were implemented on the three chosen platforms, a reduction from the fifteen kinemes of our previous work.
Of those fifteen kinemes, some were combined into one, such as \textit{Ascend} and \textit{Descend} which were combined into \textit{Indicate Movement}.
The rest were simply excluded from these experiments to enable better testing of the more important kinemes. 
The kinemes excluded include: \textit{Indicate Battery}, \textit{Possibly}, and \textit{I'm Lost}. 
These kinemes were considered less important due to their infrequent use.
Note that while \textit{Indicate Object} and \textit{Indicate Movement} are capable of indicating in multiple directions, a right and left version of each were tested for this work.

The kinemes were implemented for each platform using ROS, the Robot Operating System~\cite{quigley2009ros}. 
They are structured as a set of ROS services provided by a central package, \textit{rcvm\_core}.  
For each robot, an implementation of a server which creates the actual robot motions in response to each kineme's service call is written, forming the implementation packages \textit{rcvm\_aqua}, \textit{rcvm\_matrice}, and \textit{rcvm\_turtlebot}.

\subsection{Aqua AUV} 
The Aqua AUV is a 5 degree-of-freedom autonomous underwater vehicle, capable of fast and complex motion in $X$, $Z$, $\phi$, $\theta$, and $\psi$. 
The most maneuverable of the vehicles used in this work, Aqua is also the most difficult with which to work.
It requires access to water testing time to fine-tune the kinemes since simulators can only model the hydrodynamics and ballasting of the robot to a certain extent.
Aqua has two cameras in front, which are treated as analogous to the eyes of a head.
Pitching up and down represents a \textit{Yes} kineme, while frantic \enquote*{looking back and forth} motion is used to signal a \textit{Danger} kineme.

\subsection{Matrice 100 UAV}
The Matrice 100 is a quad-rotor aerial drone, in our case with a 3-DOF camera gimbal attached. 
Due to issues flying indoors, for this experiment the Matrice has been limited to only using its gimbal.
Kinemes have been developed using both the gimbal and the drone's flight, but for this work only the camera gimbal is in use. 
The gimbal is capable of free rotation in $\phi$, $\theta$, and $\psi$. 
In its implementation, the gimbal was treated as analogous to a single-eyed head, using pitch up and down to represent a \textit{Yes} kineme and yaw for a \textit{No}.
The required kinemes are all implemented as different combinations of gimbal rotations.

\input{figures/study_flow.tex}

\subsection{Turtlebot2 Terrestrial Robot}
The Turtlebot2 is a 3-DOF terrestrial robot, capable of free motion in $X$, $Y$, and $\theta$.
While it also has three degrees of freedom, the axes of its motion are different than the Matrice.
The Turtlebot2 also has no clear anthropomorphic features, although the Asus Xtion RGB-D sensor mounted on its platform might be seen as eyes.
Due to the Turtlebot2's inability to control its pitch, the \textit{Yes} kineme was implemented by rapidly moving the robot back and forth, which generated a head nodding effect by the rocking of the platform. 
Kinemes involving yaw, such as \textit{No} or \textit{Danger} were implemented using a combination of yaws and movement in $X$.
Interestingly, the Turtlebot2's motion is significantly more constrained than the other platforms considered, making it easier to produce dynamic motion with large changes in velocity, making the kinemes seem more \enquote{animated}.

%% file: figures/study_flow.tex
\begin{figure}
    \vspace{2mm}
    \centering
    \begin{subfigure}[b]{0.9\textwidth}
        \includegraphics[width=1\linewidth]{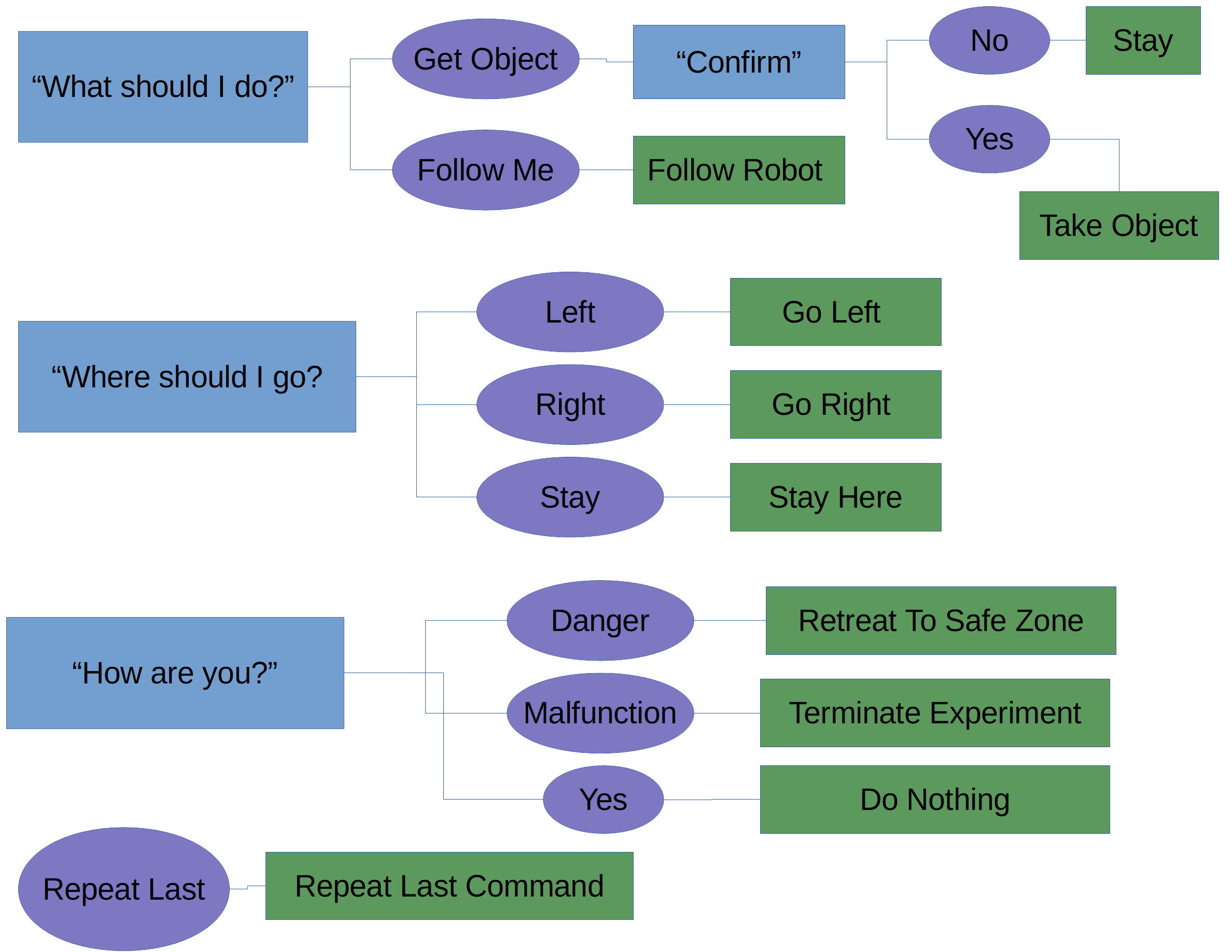}
        \caption{Flow chart depicting the interaction process.}
        \label{fig:flow_flow}
    \end{subfigure}
    \begin{subfigure}[b]{0.9\textwidth}
        \includegraphics[width=1\linewidth]{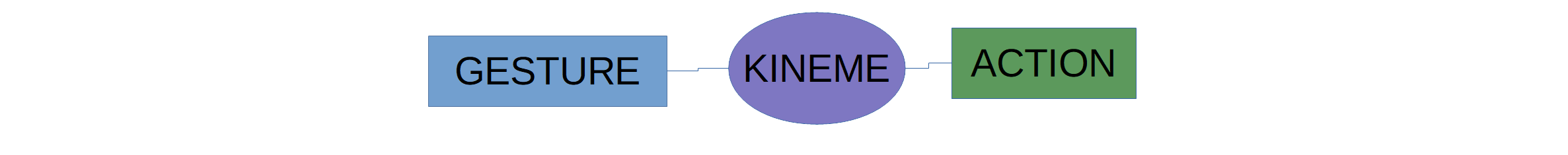}
        \caption{Key to the above flowchart.}
        \label{fig:flow_key}
    \end{subfigure}
    \caption{The above chart (a) shows the way that users will interact with the robots: a cycle of gestures, kinemes in response, followed by actions.}
    \label{fig:flow}
    \vspace{-7mm}
\end{figure}

%% file: src/experimentalsetup.tex
\section{EXPERIMENTAL SETUP}
\input{figures/study_view.tex}

To evaluate the effectiveness of kineme communication systems, we ran a small human study testing the systems on all three robots by putting participants through a set of simulated interaction scenarios. 

\subsection{Study Design}
Participants were told to ask the robot one of three questions, pay attention to the robot's response delivered via kineme, and take the appropriate action. 
A flow chart which shows the possible sequence of questions, kineme response, and actions can be found in Figure \ref{fig:flow}.
Additionally Figure \ref{fig:study_view} contains an example of the view a participant would see in an Aqua study session.
The question asked was communicated using a set of gestures based on relevant American Sign Language signs for \enquote*{What should I do?}, \enquote*{Where should I go?}, and \enquote*{How are you?}, with another sign for \enquote*{Confirm.}
The users were told that the robot would observe their gesture and automatically select a kineme to display. 
However, the kinemes were actually manually selected pseudo-randomly in secret, with consideration given to displaying each kineme close to the same number of times across the study.
This makes the study a Wizard-Of-Oz study in the context of the human-to-robot communication portion of the study, as the gesture recognition is handled by a human rather than an autonomous system.
After each interaction loop, users were asked for a confidence between one and five (five being high) on the accuracy of the whole interaction.
This confidence was recorded, along with the total time of the interaction loop and the sequence of question to kineme to action.
For each robot, participants completed an interaction session composed of between ten and fifteen interactions.

Due to physical constraints, it was only possible to schedule participants to do the Turtlebot2 and Matrice robot interactions together, followed by the Aqua interactions soon after at the University Of Minnesota Aquatic Center. 
Two participants were able to reverse the order (Aqua first, then Matrice and Turtlebot2), and the order between Matrice and Turtlebot2 was varied within all participants. 
Participants are currently being scheduled for further studies to bring the total number of participants up to 24 and cover all 6 possible orderings of robot interaction sessions.

\subsection{Study Population and Education}
The study population (N=8) was comprised of participants who were mostly age 21-34, 75\% male and 25\% female, 55.66\% Asian  and 44.44\% White or Caucasian. 
Participants self reported on their experience with robots on an ordinal scale from zero to one hundred, rating themselves at 49.20 for marine robots, 54.29 for aerial robots, and 54.00 for terrestrial robots.

All participants were provided with educational material before the study began, to be completed at their own pace. 
The educational material familiarized them with the study layout and flow (using the image in Figures \ref{fig:flow}), the gestures they would use for input, and the kinemes for all three robots, shown in a random order. 
Participants took an average of eighteen minutes to complete their education (max=$33$ minutes, min=$9$ minutes) and completed the education an average of fifteen hours before their first sessions with the robots (max=$52$ hours, min=$1$ hour). 
Users were shown Figure \ref{fig:flow} briefly before their first interaction session to refresh their memory. 

%% file: figures/study_view.tex
\begin{figure}
    \vspace{2mm}
    \centering
    \includegraphics[width=0.9\textwidth]{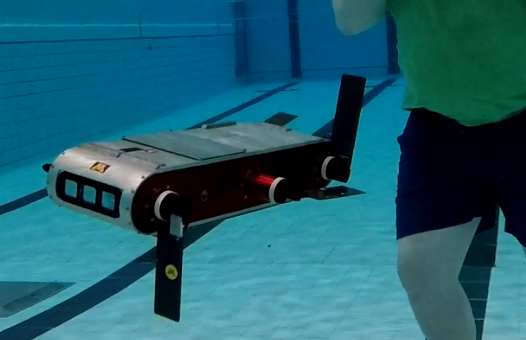}
    \caption{An image captured from the participant view of the Aqua study session.}
    \label{fig:study_view}
    \vspace{-7mm}
\end{figure}

%% file: src/results.tex
\section{RESULTS}

\subsection{Quantitative Results}

\input{figures/system_compare.tex}

\subsubsection{Accuracy}
The accuracy of the kineme systems as seen in Table \ref{tab:sys_compare} falls within acceptable levels, given the experimental nature of this work.
The kinemes are clearly not ready for systems-critical communication, but accuracy being in the $60$\%-$70$\% realm is promising, as users are recognizing the kinemes at a rate significantly higher than random chance.
The Matrice's high accuracy is likely caused by how humanoid the camera gimbal looks. 
Several participants remarked that it looked like a big eye, drawing comparisons to characters from media with similar appearances (GLaDOS, Luxo Jr., etc.).
Somewhat surprisingly, the Turtlebot2 comes second, with many participants remarking on how \enquote{cute} it was. 
The ranking of the Aqua AUV as last is likely a result of several experiment-day malfunctions.
A previously tested kineme, \textit{Follow Me}, began to experience malfunctions, achieving only 33\% accuracy. 

There are also some clear winners and losers between the individual kinemes, as seen in Table \ref{tab:kineme_comparison}. 
\textit{Repeat Last} performs very poorly, but \textit{Affirmative}, \textit{Negative}, and the \textit{Indicate Movement} kinemes perform well.  
The fact that the two directions of both \textit{Indicate Movement} and \textit{Indicate Object} have similar accuracies also indicates that participants are not having a harder time with one direction over the other.

Another promising feature of these results can be seen in the confusion matrices of Tables \ref{tab:confusion_aqua} - \ref{tab:confusion_turtlebot}.  Most kinemes clearly are being well identified as their own, with a few clear aberrations. 
The \textit{Indicate Movement} and \textit{Indicate Object} kinemes are not frequently being confused with one another, which is a positive result.
Despite both being directionally focused, the confusion between the two kinemes does not appear to be a significant problem.
However, the \textit{Indicate Object} kinemes for the Matrice, which contain a nod in the direction of an object, have been misidentified several times as the \textit{Affirmative} kineme. 
From this result, we can surmise that using one kineme in the composition of another would not be optimal. 
The development of a prefix-free set of kinemes, where each kineme starts with some unique, distinct motion to avoid any conflicts in perception, would be one way to address that concern.

\subsubsection{Efficiency and Confidence}
The average interaction time for all kineme systems is around $27$ seconds, with the Turtlebot2 achieving the fastest times, and the Aqua AUV bringing up the rear at $34.75$ seconds. 
These interaction times are obviously affected by the duration of the kinemes in question, but the ease with which participants recognize each kineme also affects the interaction times.

As far as average confidences, users have the most confidence in the Turtlebot2 interactions and the least in Aqua interactions. 
However, all systems achieve confidence over $3$, which is the halfway mark on the scale which users were given. 
The \textit{Affirmative}, \textit{Indicate Stay}, \textit{Negative}, and \textit{Indicate Movement (Right)} kinemes achieve the highest overall average confidences (all above $4$).
The \textit{Repeat Last} kineme is the only kineme with an average confidence below $3$, bringing the average down.

\input{figures/kineme_compare.tex}

\subsection{Discussion}
While the results shown in this work are clearly insufficient for an industry-level deployment of these kineme communication systems, they represent a positive step forward in the development of this communication method.
This is the first physical implementation of these kinemes, and this study revealed several prominent areas for improvement, particularly in the Aqua implementation of kinemes. 
Once these areas are addressed with further development, the accuracy and confidence for each system should rise while the interaction times fall. 

Another aspect of the results is the level of education users had.
While participants received some quick education (average of eighteen minutes), anyone with experience in using these systems would begin to perform better over time. 
This effect can be observed in the pilot studies of our previous work~\cite{fulton_robot_2018} and has been borne out by experience in this study. 
In multiple instances a user misidentified a kineme with low confidence, saw the correct kineme for that meaning later, and when shown the first kineme again correctly identified it with high confidence.

\subsection{Participant Experience}
After the completion of their study sessions, participants were given a brief survey measuring their opinions on the kineme systems for each robot. 
Overall, the Matrice kinemes were rated easiest to use, least confusing, and fastest to use, though there was concern among some participants about the effectiveness of the system at a distance greater than 10 feet. 
The Aqua kinemes were rated as the slowest and hardest to understand, and only slightly more confusing than Turtlebot2 kinemes. 
The Turtlebot2 kinemes were considered the most preferable to be used in place of a digital display or speech synthesis system, with the Matrice system tying with the option of not using any of the systems in place. 
Overall in specific questions about each kineme system, user opinions matched their quantitative results; the more accurate a user was with a kineme system, the more likely they were to rate it highly in terms of ease of use.  
Participants were moderately sure that they had made mistakes and blamed them on a combination of confusing kinemes and forgetting the correct response to a kineme.
With the exception of the Aqua kinemes, on which participants were more lukewarm, participants expressed no safety concerns with using any of the systems in an operational environment.

\subsection{Kineme Development Prescriptions}
From the experience gained in the study, we develop a set of prescriptions for future kineme constructions and improvement of current kineme systems.

\begin{itemize}
    \item Analyze motion prefix similarity of kinemes and work to separate kinemes such that their initial motions are not easily confused with each other.
    \item Make use of motion dynamics, creating kinemes with varied velocity changes to elicit greater responses.
    \item Develop a completely two-directional interaction language, to enable the use of interaction context to increase accuracy.
\end{itemize}

\input{figures/preference.tex}
\input{figures/system_confusion.tex}

%% file: figures/system_compare.tex
\begin{table}[]
    \vspace{2mm}
    \centering
    \begin{tabular}{l|c|c|c}
         System & Accuracy & Avg. Time & Avg. Conf. \\ \hline
         Aqua & 60.0\% & 34.75  & 3.01 \\
         Matrice & \textbf{76.62\%} & 24.41 &  3.83\\
         Turtlebot2 & 68.83\% & \textbf{23.4} & \textbf{3.93}\\
         \end{tabular}
    \caption{Comparison between Aqua, Matrice, and Turtlebot2 kineme systems in terms of  accuracy, average interaction time, and average confidence.}
    \label{tab:sys_compare}
\end{table}

%% file: figures/kineme_compare.tex
\begin{table}[]
    \vspace{2mm}
    \centering
    \begin{tabular}{l|c|c|c}
        Kineme &  Accuracy & Avg. Time & Avg. Conf\\ \hline
        Affirmative & \textbf{95.00}\% & 21.25 & \textbf{4.50}  \\
        Negative & 80.00\% & \textbf{20.85} & 4.05 \\
        Follow Me& 50.00\% & 30.66 & 3.31  \\
        Indicate Move (Left) & 88.46\% & 24.77 & 3.77 \\
        Indicate Move (Right)& 86.36\% & 23.18 & 4.05 \\
        Indicate Object (Left) & 57.89\% & 28.21 & 3.74 \\
        Indicate Object (Right) & 52.63\% & 35.42 & 3.16 \\
        Indicate Stay& 91.30\% & 23.78 & 4.30 \\
        Danger & 64.29\% & 32.18 & 3.21 \\
        Malfunction & 77.27\% & 29.09 & 3.50 \\
        Repeat Last& 25.00\% & 29.00 & 2.79
    \end{tabular}
    \caption{Comparison of accuracy, average interaction time and average confidence per kineme, along with counts of how many times the kineme was shown.}
    \label{tab:kineme_comparison}
    \vspace{-6mm}
\end{table}

%% file: figures/preference.tex
\begin{figure}
    \centering
    \begin{subfigure}{.9\textwidth}
        \begin{subfigure}{.3\textwidth}
            \centering
            \includegraphics[width=1\textwidth]{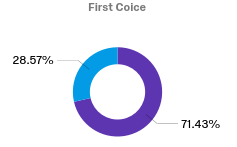}
            \caption{}
        \end{subfigure}
        \begin{subfigure}{.3\textwidth}
            \centering
            \includegraphics[width=1\textwidth]{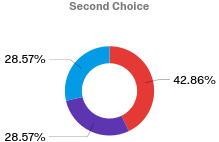}
            \caption{}
        \end{subfigure}
        \begin{subfigure}{.3\textwidth}
            \centering
            \includegraphics[width=1\textwidth]{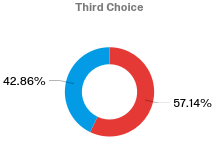}
            \caption{}
        \end{subfigure}
    \end{subfigure}
    \begin{subfigure}{.9\textwidth}
        \centering
        \includegraphics[width=1\textwidth]{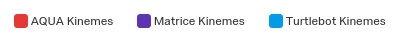}
    \end{subfigure}
    \caption{User ranking of the three kineme systems in terms of preference. (a) is users' first choice, (b) is their second, (c) is the third.}
    \label{fig:preference}
    \vspace{-7mm}
\end{figure}

%% file: figures/system_confusion.tex
\newcommand\items{11}   
\arrayrulecolor{white} 

\begin{table*}[]
    \centering
\noindent\begin{tabular}{cr*{\items}{|E}|}
\multicolumn{1}{c}{} &\multicolumn{1}{c}{} &\multicolumn{\items}{c}{\textbf{Kinemes Identified}} \\ \hhline{~*\items{|-}|}
\multicolumn{1}{c}{} & 
\multicolumn{1}{c}{} & 
\multicolumn{1}{c}{\rot{Affirmative}} & 
\multicolumn{1}{c}{\rot{Negative}} & 
\multicolumn{1}{c}{\rot{Follow Me}} & 
\multicolumn{1}{c}{\rot{Indicate Motion (Left)}} & 
\multicolumn{1}{c}{\rot{Indicate Motion (Right)}} & 
\multicolumn{1}{c}{\rot{Indicate Object (Left)}} & 
\multicolumn{1}{c}{\rot{Indicate Object (Right)}} & 
\multicolumn{1}{c}{\rot{Indicate Stay}} & 
\multicolumn{1}{c}{\rot{Danger}} & 
\multicolumn{1}{c}{\rot{Malfunction}} & 
\multicolumn{1}{c}{\rot{Repeat Last}} \\ \hhline{~*\items{|-}|}
\multirow{\items}{*}{\rotatebox{90}{\textbf{Kinemes Shown}}} 
&Affirmative & 4 & 0 & 0 & 0 & 0 & 0 & 0 & 0 & 0 & 0 & 0 \\ \hhline{~*\items{|-}|}
&Negative & 3 & 4 & 0 & 0 & 0 & 0 & 0 & 0 & 0 & 0 & 0 \\ \hhline{~*\items{|-}|}
&Follow Me & 0 & 0 & 3 & 0 & 0 & 4 & 0 & 1 & 0 & 0 & 0 \\ \hhline{~*\items{|-}|}
&Indicate Motion (Left) & 0 & 0 & 0 & 7 & 0 & 0 & 0 & 0 & 0 & 0 & 0 \\ \hhline{~*\items{|-}|}
&Indicate Motion (Right) & 0 & 0 & 2 & 0 & 3 & 0 & 0 & 0 & 0 & 0 & 0 \\ \hhline{~*\items{|-}|}
&Indicate Object (Left) & 0 & 0 & 0 & 0 & 0 & 1 & 0 & 1 & 0 & 0 & 0 \\ \hhline{~*\items{|-}|}
&Indicate Object (Right)& 0 & 0 & 1 & 0 & 0 & 0 & 5 & 0 & 1 & 0 & 0 \\ \hhline{~*\items{|-}|}
&Indicate Stay & 0 & 0 & 1 & 0 & 1 & 0 & 0 & 5 & 0 & 0 & 0 \\ \hhline{~*\items{|-}|}
&Danger & 0 & 0 & 0 & 0 & 0 & 0 & 0 & 1 & 4 & 2 & 1 \\ \hhline{~*\items{|-}|}
&Malfunction & 0 & 0 & 0 & 0 & 0 & 0 & 0 & 0 & 0 & 4 & 0 \\ \hhline{~*\items{|-}|}
&Repeat Last & 0 & 0 & 1 & 1 & 0 & 1 & 0 & 0 & 0 & 2 & 2 \\ \hhline{~*\items{|-}|}
\end{tabular}
    \caption{Confusion matrix for kineme identification for the \textbf{Aqua AUV}.}
    \label{tab:confusion_aqua}
\end{table*}

\arrayrulecolor{white} 

\begin{table*}[]
    \centering
\noindent\begin{tabular}{cr*{\items}{|E}|}
\multicolumn{1}{c}{} &\multicolumn{1}{c}{} &\multicolumn{\items}{c}{\textbf{Kinemes Identified}} \\ \hhline{~*\items{|-}|}
\multicolumn{1}{c}{} & 
\multicolumn{1}{c}{} & 
\multicolumn{1}{c}{\rot{Affirmative}} & 
\multicolumn{1}{c}{\rot{Negative}} & 
\multicolumn{1}{c}{\rot{Follow Me}} & 
\multicolumn{1}{c}{\rot{Indicate Motion (Left)}} & 
\multicolumn{1}{c}{\rot{Indicate Motion (Right)}} & 
\multicolumn{1}{c}{\rot{Indicate Object (Left)}} & 
\multicolumn{1}{c}{\rot{Indicate Object (Right)}} & 
\multicolumn{1}{c}{\rot{Indicate Stay}} & 
\multicolumn{1}{c}{\rot{Danger}} & 
\multicolumn{1}{c}{\rot{Malfunction}} & 
\multicolumn{1}{c}{\rot{Repeat Last}} \\ \hhline{~*\items{|-}|}
\multirow{\items}{*}{\rotatebox{90}{\textbf{Kinemes Shown}}} 
&Affirmative & 7 & 0 & 0 & 0 & 0 & 0 & 0 & 0 & 0 & 0 & 0 \\ \hhline{~*\items{|-}|}
&Negative & 0 & 5 & 0 & 0 & 0 & 0 & 0 & 0 & 0 & 0 & 0 \\ \hhline{~*\items{|-}|}
&Follow Me & 2 & 0 & 7 & 3 & 0 & 0 & 0 & 0 & 0 & 0 & 0 \\ \hhline{~*\items{|-}|}
&Indicate Motion (Left) & 0 & 0 & 1 & 6 & 0 & 0 & 0 & 1 & 0 & 0 & 1 \\ \hhline{~*\items{|-}|}
&Indicate Motion (Right) & 0 & 0 & 0 & 0 & 8 & 0 & 0 & 0 & 0 & 0 & 0 \\ \hhline{~*\items{|-}|}
&Indicate Object (Left) & 2 & 0 & 0 & 0 & 0 & 6 & 0 & 1 & 0 & 0 & 0 \\ \hhline{~*\items{|-}|}
&Indicate Object (Right)& 3 & 0 & 0 & 0 & 1 & 0 & 1 & 1 & 0 & 0 & 0 \\ \hhline{~*\items{|-}|}
&Indicate Stay & 0 & 0 & 0 & 0 & 0 & 0 & 0 & 8 & 0 & 0 & 0 \\ \hhline{~*\items{|-}|}
&Danger & 0 & 1 & 0 & 0 & 0 & 0 & 0 & 2 & 6 & 1 & 0 \\ \hhline{~*\items{|-}|}
&Malfunction & 0 & 0 & 0 & 0 & 0 & 0 & 0 & 0 & 1 & 8 & 0 \\ \hhline{~*\items{|-}|}
&Repeat Last & 0 & 0 & 2 & 2 & 0 & 2 & 0 & 0 & 0 & 0 & 3 \\ \hhline{~*\items{|-}|}
\end{tabular}
    \caption{Confusion matrix for kineme identification for the \textbf{Matrice}.}
    \label{tab:confusion_matrice}
\end{table*}


\arrayrulecolor{white} 

\begin{table*}[]
    \centering
\noindent\begin{tabular}{cr*{\items}{|E}|}
\multicolumn{1}{c}{} &\multicolumn{1}{c}{} &\multicolumn{\items}{c}{\textbf{Kinemes Identified}} \\ \hhline{~*\items{|-}|}
\multicolumn{1}{c}{} & 
\multicolumn{1}{c}{} & 
\multicolumn{1}{c}{\rot{Affirmative}} & 
\multicolumn{1}{c}{\rot{Negative}} & 
\multicolumn{1}{c}{\rot{Follow Me}} & 
\multicolumn{1}{c}{\rot{Indicate Motion (Left)}} & 
\multicolumn{1}{c}{\rot{Indicate Motion (Right)}} & 
\multicolumn{1}{c}{\rot{Indicate Object (Left)}} & 
\multicolumn{1}{c}{\rot{Indicate Object (Right)}} & 
\multicolumn{1}{c}{\rot{Indicate Stay}} & 
\multicolumn{1}{c}{\rot{Danger}} & 
\multicolumn{1}{c}{\rot{Malfunction}} & 
\multicolumn{1}{c}{\rot{Repeat Last}} \\ \hhline{~*\items{|-}|}
\multirow{\items}{*}{\rotatebox{90}{\textbf{Kinemes Shown}}} 
&Affirmative & 8 & 0 & 0 & 0 & 0 & 0 & 0 & 0 & 0 & 1 & 0 \\ \hhline{~*\items{|-}|}
&Negative & 0 & 7 & 0 & 0 & 0 & 0 & 0 & 0 & 0 & 0 & 0 \\ \hhline{~*\items{|-}|}
&Follow Me & 1 & 0 & 7 & 0 & 2 & 0 & 1 & 0 & 0 & 0 & 0 \\ \hhline{~*\items{|-}|}
&Indicate Motion (Left) & 0 & 0 & 1 & 9 & 0 & 0 & 0 & 0 & 0 & 0 & 0 \\ \hhline{~*\items{|-}|}
&Indicate Motion (Right) & 0 & 0 & 1 & 0 & 8 & 0 & 0 & 0 & 0 & 0 & 0 \\ \hhline{~*\items{|-}|}
&Indicate Object (Left) & 2 & 0 & 0 & 0 & 0 & 4 & 0 & 2 & 0 & 0 & 0  \\ \hhline{~*\items{|-}|}
&Indicate Object (Right)& 1 & 0 & 0 & 0 & 0 & 0 & 4 & 0 & 0 & 0 & 0 \\ \hhline{~*\items{|-}|}
&Indicate Stay & 0 & 0 & 0 & 0 & 0 & 0 & 0 & 9 & 0 & 0 & 0 \\ \hhline{~*\items{|-}|}
&Danger & 0 & 0 & 0 & 0 & 0 & 0 & 0 & 0 & 8 & 1 & 0 \\ \hhline{~*\items{|-}|}
&Malfunction & 1 & 0 & 0 & 0 & 0 & 0 & 0 & 0 & 1 & 5 & 1 \\ \hhline{~*\items{|-}|}
&Repeat Last & 0 & 0 & 0 & 0 & 0 & 0 & 3 & 3 & 0 & 0 & 1 \\ \hhline{~*\items{|-}|}
\end{tabular}
    \caption{Confusion matrix for kineme identification for the \textbf{Turtlebot2.}}
    \label{tab:confusion_turtlebot}
\end{table*}

%% file: src/conclusion.tex
\section{CONCLUSION}
In this work we presented, for the first time, an implementation of our previously proposed kineme system for motion-based robot-to-human communication. 
This implementation supports not one but three different robots in the disparate domains of underwater, aerial, and terrestrial robotics.
The implementation leverages the anthropomorphism often applied to robots, treating cameras as eyes and using the spatial aspects of the concepts being communicated to design effective motion.
In a small user study we examined these kineme systems in terms of accuracy, average interaction times, and average confidence of interaction success.
Our results from that study were promising and provided us with a set of directions to guide our future development.

\subsection*{Future Work}
Our future work will focus on expanding the results of this study and further exploring the effects of different conditions on kineme communication. 
First and foremost, this study is still in progress. 
Data from a full twenty-four participants is planned to be collected to strengthen the findings of this work. 
Secondly, this study is treated as a pilot study for a larger study to be conducted via Amazon Mechanical Turk in the coming months.  
This study did not test for differences in interaction success at different distances or angles in comparison to the robot, or compare the kineme system to other communication systems. 
The online study will do so. 
Furthermore, this study was only a Wizard-of-Oz study, containing no implementation of a human-to-robot communication system. 
Future work will involve the implementation of such a system as well as perception algorithms to help ground the interactions by the gaze direction of the human interactant, as well as their current activity state (busy, waiting for information, speaking, etc.)

Once we have further explored the different aspects of kineme communication, there are a number of possible extensions to the kineme system that go beyond motion that are interesting, including the integration of light and sound cues into the system. 
This will undoubtedly be a feature of long-term research efforts in this area, as no single communication vector will be paramount in every domain and situation.

%% file: src/acknowledgment.tex
\section*{ACKNOWLEDGMENTS}
The authors wish to thank all study participants for their time,  Dr. Daniel Keefe for his input on RCVM in the early phases of prototyping, and Sophie Fulton for her insight on kineme design.
Special thanks to Jungseok Hong and Youya Xia, who helped to perform data entry for some study sessions.
Thanks also to Hope Mills, lifeguard for the pool study session and Marc Ho for his help in organizing pool timing.